\pdfoutput=1

\documentclass[11pt]{article}

\usepackage[final]{acl}

\usepackage{times}
\usepackage{latexsym}

\usepackage[T1]{fontenc}

\usepackage[utf8]{inputenc}

\usepackage{microtype}

\usepackage{inconsolata}

\usepackage{graphicx}

\usepackage{amsmath}
\usepackage{multirow}
\usepackage{booktabs}
\usepackage{xspace}
\usepackage{amssymb}
\usepackage{xcolor}
\usepackage{colortbl}
\usepackage{tcolorbox}

\usepackage{float}
\usepackage{algorithm}
\usepackage{algpseudocode}
\usepackage{fontawesome5}

\newcommand{\ours}{\textcolor{black}{\texttt{COM-BOM}}\xspace} %
\definecolor{lightorange}{RGB}{246 180 143}
\definecolor{lightred}{RGB}{173, 23, 89}

\newcommand{\bz}{\mathbf{z}}

\newcommand{\obsdata}{\mathcal{D}}
\newcommand{\Pfront}{\mathcal{P}}

\title{\ours: Bayesian Exemplar Search for Efficiently\\ Exploring the  Accuracy-Calibration Pareto Frontier}

\author{Gaoxiang Luo \\
  University of Minnesota \\
  \texttt{luo00042@umn.edu} \\\And
  Aryan Deshwal \\
  University of Minnesota \\
  \texttt{adeshwal@umn.edu} \\}

\begin{document}
\maketitle
\begin{abstract}

Selecting an optimal set of exemplars is critical for good performance of in-context learning. However, prior exemplar search methods narrowly optimize for predictive accuracy, critically neglecting model calibration—a key determinant of trustworthiness and safe deployment.
In this paper, we formulate exemplar selection as a multi-objective optimization problem, explicitly targeting both the maximization of predictive accuracy and the minimization of expected calibration error. 
We solve this problem with a sample-efficient Combinatorial Bayesian Optimization algorithm (\ours) to find the Pareto front that optimally trades off the two objectives of accuracy and calibration. We evaluate \ours on multiple tasks from unsaturated MMLU-Pro benchmark and find that \ours beats or matches the baselines at jointly optimizing the two objectives, while requiring a minimal number of LLM API calls.
\end{abstract}
\begin{center}
    \vspace{-1em}
    \faGithub \hspace{2pt} \href{https://github.com/GaoxiangLuo/COM-BOM}{github.com/GaoxiangLuo/COM-BOM}
\end{center}

\section{Introduction}
In‑context learning (ICL) has emerged as a powerful paradigm, enabling large language models (LLMs) to solve new tasks by conditioning on a prompt containing an instruction and a set of exemplars. While a large body of work focuses on instruction optimization, recent findings \cite{wan2024teach, ajith2023instructeval} reveal that exemplar selection contributes substantially more to ICL's performance than instructions alone. Despite this critical importance, principled approaches to exemplar selection remain surprisingly under-explored, particularly concerning the two objectives of predictive accuracy and model reliability.

\begin{figure}[t]
    \centering
    \includegraphics[width=1.0\linewidth]{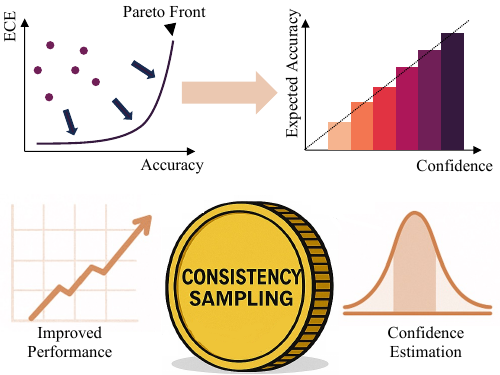}
    \caption{Optimizing for accuracy and calibration error leads to better reliability (top). Two sides of the same coin for self-consistency sampling  (bottom).}
    \label{fig:teaser}
    \vspace{-1.5em}
\end{figure}

The paper addresses this gap by re-evaluating the core goals of exemplar selection in ICL.  Prior methods for exemplar selection predominantly pursue a single objective: maximize predictive accuracy. While accuracy is undoubtedly important, many high-stakes real-world applications (such as finance, healthcare, and legal settings) require not only high accuracy but also {\em well-calibrated} confidence. Poorly calibrated models may exhibit overconfidence in erroneous predictions or under confidence in correct ones, diminishing their practical value.  Indeed, current exemplar selection strategies, by largely ignoring calibration, risk scenarios where gains in accuracy are achieved at the expense of degradation in calibration \cite{zhang2023study}. 

Consequently, in this work, we propose that exemplar selection for LLMs should be reframed as a multi-objective optimization problem where the goal is to jointly optimize for both accuracy and calibration by identifying a set of exemplars that optimally trade off these two conflicting objectives. However, our multi-objective optimization formulation of exemplar selection in ICL presents several technical challenges: 1) combinatorial nature of the search space in selecting an optimal subset of exemplars, 2) the need to identify a Pareto frontier of solutions  that represent varying trade-offs between accuracy and calibration, and 3) the expensive and noisy nature of evaluating these metrics, which often involve multiple LLM API calls. 

In order to address these challenges, we propose a principled Bayesian optimization algorithm titled: Combinatorial Bayesian Optimization of Multiple ICL Metrics (\ours). The key idea is to leverage probabilistic surrogate models defined over combinatorial inputs with multi-objective acquisition functions to intelligently guide the search towards the pareto-optimal frontier of accuracy and calibration in a sample-efficient manner i.e. minimizing the number of calls to the expensive LLM API calls. The key contributions of this paper are: 
\begin{enumerate}
    \item We introduce a new formulation of exemplar selection in ICL as multi‑objective optimization problem, explicitly targeting two important metrics of accuracy and calibration.
    \item We propose \ours, a combinatorial multi-objective Bayesian optimization algorithm, specifically designed to tackle the key technical challenges of multiple objectives, combinatorial search spaces, and expensive and noisy evaluations specific to this problem.
    \item We empirically validate our approach, showcasing its ability to identify exemplar sets that achieve better Pareto frontier of the two objectives compared to existing methods, while requiring fewer LLM evaluations.
\end{enumerate}

\section{Background and Related Work}

\paragraph{Exemplar Selection.}
The simplest form of methods for exemplar selection in ICL use existing lexical or embedding-based similarity metrics to retrieve examples most similar to the test input from a pool of candidates \cite{wang2023learning}. Another line of work trains bespoke retrievers specifically to pick in-context examples \cite{rubin2021learning, cheng2023uprise}. One drawback of these methods is the amount of redundancy in their selected exemplars. Diversity based approaches aim to overcome this drawback by encouraging diversity in the selected exemplars \cite{ye2023compositional}. However, all of these approaches focus on predictive accuracy as the primary performance metric. Recent works demonstrate that exemplar selection contributes more to ICL's accuracy compared to instruction optimization~\cite{wan2024teach, ajith2023instructeval}. Recently, \citet{wan2025from-582} proposed BRIDGE algorithm for exemplar selection in the many-shot ICL setting which alternates between generating and optimizing a set of exemplars. The optimization step in BRIDGE leverages Bayesian optimization to solve a bi-objective (accuracy and sparsity) problem by converting it to a single objective via random scalarization. Our approach differs from BRIDGE in that we focus on optimizing calibration and accuracy as the two objectives and directly reason about the Pareto frontier using a hypervolume based multiobjective acquisition function.

\paragraph{Multi-Objective Optimization.} 
Multi-objective optimization (MOO) \cite{emmerich2018tutorial,belakaria2019max, daulton2020differentiable} addresses problems where several, often conflicting, objective functions must be optimized simultaneously. Unlike single-objective problems that typically seek a single optimal solution, MOO aims to identify a set of solutions representing the best possible trade-offs, known as the {\em Pareto optimal set.} A solution is considered Pareto optimal if no objective function can be improved without degrading at least one other objective. The values of this set in the objective space forms the {\em Pareto front}, which characterizes the optimal attainable trade-off between the competing objectives. The core challenge in MOO lies in efficiently exploring the input space to discover or approximate this Pareto front.

\section{Methodology}
\subsection{Problem Definition: Multiobjective Formulation for Exemplar Selection \label{subsection:problem}}

In this work, we study the problem of exemplar selection in ICL where our objective is to identify an optimal subset of exemplars from a given pool $\mathcal{E} = \{e_1, e_2 ... e_m\}$ that maximizes the LLM's performance on a given task. We formalize this as a combinatorial search problem where the search space is defined over binary indicator vectors $\mathbf{z} \in \{0,1\}^m$. Each element $\bz_i$ indicates whether exemplar $e_i$ is included in the prompt or not.

While most existing work on exemplar selection focuses solely on optimizing {\em predictive accuracy}  as performance metric, many real-world applications such as clinical question answering \cite{agrawal2022large} or legal decision making \cite{lai2024large}, often demand not just correctness but also reliable model confidence. The reliability of model confidence is quantified by expected calibration error (ECE) which measures how much its confidence, i.e. the predicted probability of correctness, diverges from its accuracy, i.e. the empirical probability of correctness. In this paper, we posit that exemplar selection should be formalized as a multi-objective optimization problem that seeks to jointly optimizes for high predictive accuracy and low calibration error. Accordingly, we define LLM performance using two (often conflicting) objectives:

$\blacktriangleright$ {\bf Objective 1: Maximizing Predictive Accuracy $(f_{\text{acc}}(\bz))$} :The ability of the LLM, conditioned on the exemplar set $\bz$, to generate correct outputs. 

$\blacktriangleright$  \textbf{Objective 2: Minimizing Expected Calibration Error ($f_{\text{ECE}}(\bz)$)}: A metric to estimate  LLM's miscalibration based on its confidence (derived from its output distribution conditioned on $\bz$) and its observed accuracy.
\vspace{1ex}

{\em \noindent Goal:} Our overall problem setup in this paper is succinctly formalized as the following multi-objective optimization problem:
\begin{align}
    \max_{\bz \in \{0,1\}^m} \quad (f_{\text{acc}}(\bz), -f_{\text{ECE}}(\bz)) \label{obj:main}
\end{align}
i.e., we seek to find a set of pareto-optimal exemplar sets $\bz^*$ that offer the best possible accuracy-calibration trade-off. Please note $f_{\text{ECE}}$ is negated to maintain a consistent maximization formulation.

\subsubsection{Evaluation of Accuracy and Calibration Error Objectives \label{sec:eval_obj}}

To evaluate the effectiveness of a chosen exemplar set $\bz$ according to our multi-objective formulation (\ref{obj:main}), we need robust methods for estimating predictive accuracy ($f_{\text{acc}}(\bz)$) and ECE ($f_{\text{ECE}}(\bz)$). Our evaluation procedure is designed for black-box LLM access and leverages the key principle of {\em semantic consistency} \cite{farquhar2024detecting, zhong2022non} from multiple generated outputs.
Interestingly, most popular techniques for estimating predictive accuracy, such as self-consistency decoding methods \cite{wang2022self}, already rely on generating multiple output samples from the LLM. We describe below that ECE can be cheaply computed from these same set of samples already being generated from the LLM, an opportunity overlooked by existing exemplar selection methods. The evaluation process for a given exemplar set $\bz$ assumes a small validation set $\mathcal{D}_{\text{val}}$ and involves the following steps, logically grouped into {\em per-instance estimation} and {\em aggregate metric calculation}:

    \paragraph{Per-Instance Confidence, and Accuracy Estimation}
    For each input query $x$ (with ground truth $y$) from $\mathcal{D}_{\text{val}}$, and the chosen exemplar set $\bz$:
    \begin{enumerate}
        \item[(a)] \textit{Diverse Output Generation:} We construct a $\texttt{Prompt}[x, \bz]$ and prompt the LLM to generate $M$ diverse output sequences (potential answers) $S_{x,\bz} = \{s_1, s_2, ..., s_M\}$.

        \item[(b)] \textit{Semantic Clustering and Answer Determination:} The generated samples $S_{x,\bz}$ are clustered based on semantic equivalence \cite{farquhar2024detecting}. Sequences $s_i$ and $s_j$ are grouped into the same semantic cluster $c$ if they convey the same core meaning (e.g., via bidirectional entailment), despite lexical variations. This results in a set of semantic clusters $\mathcal{C}_{x,\bz} = \{c_1, c_2, ..., c_L\}$ (where $L \le M$). The predicted answer, $\text{ans}(x, \bz)$, is taken from the largest (most frequent) semantic cluster, $c^* = \arg \max_{c_l \in \mathcal{C}_{x,\bz}} |c_l|$.

        \item[(c)] \textit{Confidence Estimation:} The confidence in the prediction $\text{ans}(x, \bz)$ is defined as the probability of its corresponding semantic cluster $c^*$:
        $\text{conf}(x, \bz) = \max_{c_i \in \mathcal{C}_{x,\bz}} \frac{|c_i|}{M}.
        \label{eq:confidence_revised}$
        \noindent This score reflects the LLM's internal consistency in generating the chosen semantic output.
        
        \item[(d)] \textit{Per-Instance Accuracy:} The accuracy for this single input $x$ given $\bz$, denoted $\text{acc}(x, \bz)$, is 1 if $\text{ans}(x, \bz)$ matches the ground truth label $y$, and 0 otherwise.
    \end{enumerate}

    \paragraph{Aggregate Metric Calculation over the Validation Set}
    After processing all instances in $\mathcal{D}_{\text{val}}$:
    \begin{enumerate}
        \item[(a)] \textit{Overall Predictive Accuracy} ($f_{\text{acc}}(\bz)$): The overall predictive accuracy for the exemplar set $\bz$ is the average of the per-instance accuracies:
        \begin{equation}
        f_{\text{acc}}(\bz) = \frac{1}{|\mathcal{D}_{\text{val}}|} \sum_{(x,y) \in \mathcal{D}_{\text{val}}} \text{acc}(x, \bz)
        \label{eq:accuracy_revised}
        \end{equation}

        \item[(b)] \textit{Expected Calibration Error} ($f_{\text{ECE}}(\bz)$): ECE \cite{naeini2015obtaining} quantifies the mismatch between the model's confidence and its empirical accuracy. It is computed as follows:
        \begin{enumerate}
            \item[(i)] Collect all $(\text{conf}(x, \bz), \text{acc}(x, \bz))$ pairs for $(x,y) \in \mathcal{D}_{\text{val}}$.
            \item[(ii)] Divide the confidence interval $[0, 1]$ into $K$ equally spaced bins.
            \item[(iii)] For each bin $k$ ($1 \leq k \leq K$):
            \begin{itemize}
                \item Let $\mathcal{B}_k$ be the set of input-output pairs $(x,y)$ whose $\text{conf}(x, \bz)$ falls into bin $k$.
                \item Calculate the average confidence in bin $k$: 
                 $ \text{conf}_{\mathcal{B}_k} = \frac{1}{|\mathcal{B}_k|} \sum_{(x,y) \in \mathcal{B}_k} \text{conf}(x, \bz). $
                \item Calculate the average accuracy in bin $k$: 
                $\text{acc}_{\mathcal{B}_k} = \frac{1}{|\mathcal{B}_k|} \sum_{(x,y) \in \mathcal{B}_k} \text{acc}(x, \bz).$
            \end{itemize}
            \item[(iv)] The ECE for exemplar set $\bz$ is the weighted average of the absolute differences between average accuracy and average confidence across all bins:
            {\fontsize{10pt}{8.5pt}
            \begin{equation}
            f_{\text{ECE}}(\bz) = \sum_{k=1}^{K} {\small \frac{|\mathcal{B}_k|}{|\mathcal{D}_{\text{val}}|}} \cdot |\text{acc}_{\mathcal{B}_k} - \text{conf}_{\mathcal{B}_k}|
            \label{eq:ece_revised}
            \end{equation}
            }
        \end{enumerate}
        A lower $f_{\text{ECE}}(\bz)$ indicates better calibration, meaning the LLM's confidence $\text{conf}(x, \bz)$ more accurately reflects its likelihood of being correct.  This evaluation methodology works with only black-box access to API LLMs.
    \end{enumerate}

\subsubsection{Key Optimization Challenges}  There are two technical challenges that arise because of our evaluation procedure.  
{\em First,} each objective evaluation of the objective function is black-box and expensive (both monetary costs of LLM API calls and latency costs of multiple samples) which necessitates {\em sample-efficient} methods that can find high-quality exemplar sets with a minimal number of objective function evaluations. {\em Second,} each objective evaluation is black-box noisy estimate of the true ground truth. Therefore, we want to develop black-box optimization algorithms that are robust to noisy observations. 
In the subsequent sections, we detail our proposed Combinatorial Bayesian Optimization of Multiple ICL Metrics \ours algorithm to address these challenges.

\begin{figure}[b]
    \centering
    \includegraphics[width=1.0\linewidth]{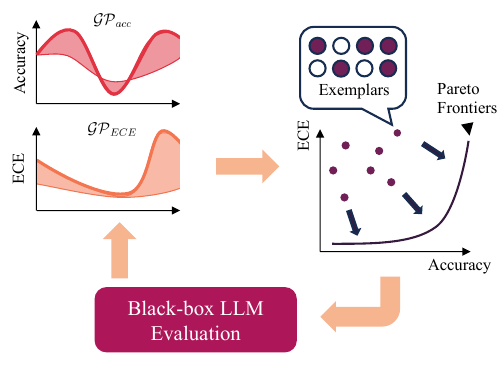}
    \caption{The BO loop with a single-task GP for each objective and multi-objective acquisition function.}
    \label{fig:bo}

\end{figure}

\begin{algorithm}[h!]
\begin{algorithmic}[1]
  \Require Exemplar pool $\mathcal{E}$, validation set $D_{val}$, evaluation budget $t_{\max}$

  \State Randomly sample $t_0$ initial $\bz$'s and evaluate the two objective functions to initialize $D_{t_0}, \Pfront_{t_0}$;
   
 \Statex $D_{t_0} = \bigl\{(\mathbf z_i, o_i)\bigr\}_{i=1}^{t_0},\quad o_i\in\{f_{\rm acc},\,f_{\rm ECE}\},$ 
  \Statex $\Pfront_{t_0}$:  $\mathrm{Pareto\, Frontier} (D_{t_0})$
  
  \For{$n = t_0 $ \textbf{to} $t_{\max}$}
    \State Fit Gaussian Process surrogate models $\tilde{f}_{\text{acc}}$, $\tilde{f}_{\text{ECE}}$ on $D_n$ \Comment{Section \ref{sec:surrogate}}
    \State Optimize the multiobjective acquisition function $ \alpha\bigl(\bz \mid D_n, \Pfront_n \bigr)$ to get next point to evaluate $\bz_{n+1}$ \Comment{Section \ref{sec:af_def}, \ref{sec:afo}}
      
    \State Evaluate objectives on $\bz_{n+1}$ by generating samples from LLM to get $o_{n+1} = (f_{\text{acc}} (\bz),\,f_{\text{ECE}} (\bz))$ \Comment{Section \ref{sec:eval_obj}}
    \State 
     $D_{n+1}$ =  $D_n \;\cup\!\{(\bz_{n+1},(o_{n+1}))\}$
     \State $\Pfront_{n+1} \!\leftarrow\!\mathrm{Pareto\, Frontier}(\obsdata_{n+1})$
  \EndFor
  \Ensure Approximate Pareto frontier $\Pfront_{t_{\max}}$ and corresponding Pareto set  of the two objectives: accuracy and expected calibration error.
\end{algorithmic}

\caption{Pseudocode for \ours}
\label{alg:com-bom-exemplar}
\end{algorithm}

\subsection{Bayesian Optimization Solution: \ours}

Bayesian optimization (BO) \cite{garnett2023bayesian} is an effective and principled framework for black-box optimization, particularly when objective function evaluations are expensive, as is the case in our problem setting where each evaluation requires multiple LLM calls for a given input $\bz$ (exemplar set). The key idea in BO is to construct a probabilistic surrogate model and using it to guide sequential function evaluations. Our approach \ours instantiates the Bayesian optimization framework with its three core components, each tailored to address the unique challenges of our problem setup. 

{\em First}, two Gaussian process surrogate models  are constructed, which represents our belief about how different exemplar set choices ($\bz$) influence both predictive accuracy ${f}_{acc}$ and expected calibration error ${f}_{\text{ECE}}$. {\em Second}, a multi-objective acquisition function is employed which quantifies the utility of evaluating the objective at different exemplar sets by balancing exploitation of regions where the surrogate predicts high function values against exploration of regions with high predictive uncertainty.  
{\em Third}, an acquisition function optimization procedure is used to efficiently identify the most promising exemplar set to evaluate next, thereby guiding the iterative search process towards optimal inputs with minimum objective function calls. We describe the details of each component in the subsequent sections. Our  approach is implemented in the BoTorch library \cite{balandat2020botorch}. 

\subsubsection{Gaussian Process Surrogate Models with Exponentiated Hamming Kernel \label{sec:surrogate}}

Gaussian Processes (GPs) \cite{williams1995gaussian} are an effective class of probabilistic  model for surrogate modeling in Bayesian optimization due to their ability to provide principled uncertainty quantification. GP models are mainly characterized by the choice of the kernel or covariance function ($k_H(\mathbf z,\mathbf z')$) and a mean function ($\mu(\mathbf z)$; parametrized as a learnable constant).  We use an independent Gaussian-process model, one for each objective, with an exponentiated hamming distance kernel \cite{wan2021think} which is suitable for combinatorial space of exemplar selection, i.e., 
\[
\tilde{f}_{\text{acc}}, \tilde{f}_{\text{ECE}} \sim \mathcal{GP}\bigl(\mu(\mathbf z),\,k_H(\mathbf z,\mathbf z')\bigr)
\]
with kernel/covariance function
\begin{align}
k_H(\mathbf z,\mathbf z')
=\exp\left(-\,
d_{\text{H}}(\mathbf z,\mathbf z')\right), \\
d_{\text{H}}(\mathbf z,\mathbf z') = \sum_{j=1}^{m} \frac{1}{\ell_j} \cdot \mathbf{1}[z_j \ne z'_j].
\end{align}
where ${\ell_j}$ is a separate lengthscale parameter for each input dimension, typically referred as automatic relevance determination. These dimension-specific lengthscales allow the GP to learn the relative importance of different input dimensions. 

Given observations:
\[
D_n = \bigl\{(\mathbf z_i, o_i)\bigr\}_{i=1}^n,\quad o_i\in\{f_{\rm acc},\,f_{\rm ECE}\}, 
\]
define the kernel matrix and cross‐covariances
\[
[K_H]_{ij} = k_H(\mathbf z_i,\mathbf z_j), 
\quad 
\mathbf k_H(\mathbf z) = \bigl[k_H(\mathbf z,\mathbf z_i)\bigr]_{i=1}^n.
\]
Then for each objective $\tilde f\in\{\tilde f_{\rm acc},\tilde f_{\rm ECE}\}$, the posterior predictive at a new point $\mathbf z$ is computed in closed form, providing both a mean prediction and uncertainty estimate:
\[
p\bigl(\tilde f(\mathbf z)\mid \mathbf z, D_n\bigr)
= \mathcal{N}\bigl(\mu_n(\mathbf z),\,s_n^2(\mathbf z)\bigr), \text{with}
\]
\[
\mu_n(\mathbf z)
= \mu(\mathbf z)
+ \mathbf k_H(\mathbf z)^\top
\bigl[K_H + \sigma_n^2 I\bigr]^{-1}
\bigl(\mathbf o - \mu(\mathbf Z)\bigr),
\]
\[
s_n^2(\mathbf z)
= k_H(\mathbf z,\mathbf z)
- \mathbf k_H(\mathbf z)^\top
\bigl[K_H + \sigma_n^2 I\bigr]^{-1}
\mathbf k_H(\mathbf z),
\] 
where $\sigma_n^2$ is the observation noise variance, $\mathbf o=[o_i]_{i=1}^n$, and $\mu(\mathbf Z)=[\mu(\mathbf z_i)]_{i=1}^n\,$.

{\em Notation Remark:} As mentioned in problem definition \ref{subsection:problem}, our goal is to {\em minimize} ECE and {\em maximize} accuracy. 
In practice, we negate the ECE observations to transform the problem into a consistent maximization formulation for both  objectives. Thus, our approach's implementation operates on $\{f_{\rm acc},\, -f_{\rm ECE}\}$.  However, in our discussion, we refer to ECE in its original, unnegated form ($f_{\rm ECE}$) where lower values are better.

\subsubsection{Hypervolume based Multi-objective Acquisition Function \label{sec:af_def}}

The acquisition function quantifies the utility of evaluating a candidate exemplar set $\bz$ based on the current probabilistic models of the objectives ($p\bigl(\tilde f(\mathbf z)\mid \mathbf z, D_n\bigr)$). In multi-objective setting, an effective acquisition function must jointly reason about the two objectives and guide the search towards improving the incumbent Pareto frontier $\mathcal{P}_n$. 

\begin{figure}
    \centering
    \includegraphics[width=1.0\linewidth]{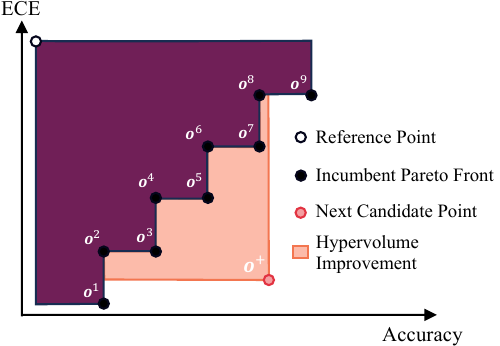}
    \caption{Illustration of Hypervolume improvement acquisition function for a candidate point (Section \ref{sec:af_def}).}
    \label{fig:ehvi}
    \vspace{-1.5em}
\end{figure}

One such effective choice for a multiobjective acquisition function is the Expected  Improvement in the Hypervolume Indicator (EHVI)  \cite{daulton2020differentiable}. The hypervolume itself is the Lebesgue measure of the objective space region dominated by a given Pareto frontier and bounded by a reference point \cite{emmerich2018tutorial}. Intuitively, for a set of non-dominated solutions (the Pareto frontier), the hypervolume quantifies the coverage of the objective space that these solutions collectively achieve i.e. gives us a global view of the solution space that is typically missed by the naive baseline of scalarizing the two objectives. 
EHVI, therefore, measures the expected increase in this dominated hypervolume if a new candidate $\bz$ were to be evaluated and added to the set of known solutions (see Figure \ref{fig:ehvi} for illustration). Maximizing EHVI aims to expand the coverage and quality of the approximated Pareto front.

Let \(\mathcal P_n=\{\mathbf o^i\}_{i=1}^{|\mathcal P_n|}\) be the current  Pareto frontier and \(HV(\cdot)\) its hypervolume relative to a fixed reference point.  Then the Expected Hypervolume Improvement is defined as:\\
{\fontsize{10pt}{8.5pt}
\begin{equation}
\begin{split}
\alpha(\mathbf z \mid D_n, \mathcal P_n)
=\mathbb E_{\mathbf F(\mathbf z)\mid D_n}\Bigl[HV\bigl(\mathcal P_n\cup\{\mathbf F(\mathbf z)\}\bigr)\;  -\; \\ 
HV(\mathcal P_n)\Bigr]
\end{split}
\end{equation}
}

\begin{align*}
\mathbf F(\mathbf z) \;=\;
\begin{pmatrix}
\tilde f_{\rm acc}(\mathbf z) \\[4pt]
\tilde f_{\rm ECE}(\mathbf z)
\end{pmatrix}
\sim
\mathcal N\Bigl(\,
\boldsymbol\mu_n(\mathbf z),\,
\boldsymbol\Sigma_n(\mathbf z)
\Bigr)
\end{align*}

\[
\boldsymbol\mu_n(\mathbf z)
=
\begin{pmatrix}
\mu_{n}^{\rm acc}(\mathbf z) \\[4pt]
\mu_{n}^{\rm ECE}(\mathbf z)
\end{pmatrix},
\]
\[
\boldsymbol\Sigma_n(\mathbf z)
=
\operatorname{diag}\bigl(s_{n,{\rm acc}}^2(\mathbf z),\,s_{n,{\rm ECE}}^2(\mathbf z)\bigr)
\]
where each mean and variance is given by the GP posterior.  
In this work, we utilize the related Noisy Expected Hypervolume Improvement (NEHVI) acquisition function, as introduced by  \cite{daulton2021parallel},  which is suitable for settings with noisy observations. NEHVI extends EHVI by marginalizing out the uncertainty over the true Pareto frontier given the history of noisy evaluations. 

\subsubsection{Acquisition Function Optimization with Greedy Hill-Climbing \label{sec:afo}}
Following prior work in Bayesian optimization, we pick the next candidate exemplar set $z_{t+1}$ by optimizing the acquisition function over the combinatorial space $\{0, 1\}^m$ using a trust-region based \cite{eriksson2019scalable, wan2021think} heuristic local search strategy, which we found to work well in practice.  The key idea is to optimize the acquisition function using a local search based greedy hill-climbing procedure within a trust region defined as a Hamming ball centered around a promising candidate $z_{center}$. In our work,  the center point $z_{center}$ is selected  by identifying the point in the non-dominated set that contributes most to the hypervolume of the current approximate Pareto front \cite{daulton2022multi, oh2019combinatorial}. Multiple random restarts are used to improve the performance of the greedy hill-climbing  optimizer.  

\section{Experiments}

\begin{table*}[htbp]
\centering
\small
\begin{tabular}{l|cc|cc|cc|cc|cc}
\toprule
& \multicolumn{2}{c|}{No Exemplars} & \multicolumn{2}{c|}{All Exemplars} & \multicolumn{2}{c|}{$\text{Nearest}_{k=10}$} & \multicolumn{2}{c|}{$\text{Diversity}_{k=10}$} & \multicolumn{2}{c}{Ours} \\
\cmidrule(lr){2-3} \cmidrule(lr){4-5} \cmidrule(lr){6-7} \cmidrule(lr){8-9} \cmidrule(lr){10-11}
Domain & Acc & ECE & Acc & ECE & Acc & ECE & Acc & ECE & Acc & ECE \\
\midrule
biology & 77.19 & 19.10 & 79.24 & 18.40 & 76.60 & 20.29 & 76.90 & 24.32 & \cellcolor{lightred!15}79.24 & \cellcolor{lightorange!15}18.37 \\
business & 40.74 & 32.97 & \cellcolor{lightred!15}41.27 & 27.25 & 39.68 & 30.72 & 38.89 & 33.00 & 41.01 & \cellcolor{lightorange!15}21.54 \\
chemistry & 34.79 & 37.45 & \cellcolor{lightred!15}39.34 & 29.50 & 37.34 & 31.79 & 36.61 & 33.26 & 37.34 & \cellcolor{lightorange!15}28.38 \\
computer science & 48.94 & 35.84 & 50.53 & 32.42 & 48.94 & 37.16 & 51.06 & 35.05 & \cellcolor{lightred!15}55.85 & \cellcolor{lightorange!15}32.08 \\
economics & \cellcolor{lightred!15}63.95 & \cellcolor{lightorange!15}30.24 & 62.96 & 30.89 & 61.48 & 30.98 & 62.47 & 33.20 & 63.70 & 30.74 \\
engineering & 44.44 & 30.67 & 46.15 & 21.13 & \cellcolor{lightred!15}47.01 & 21.00 & 42.95 & 23.38 & \cellcolor{lightred!15}47.01 & \cellcolor{lightorange!15}19.32 \\
health & 60.20 & 33.91 & 60.20 & 32.01 & 56.12 & 32.55 & 60.71 & 31.93 & \cellcolor{lightred!15}62.24 & \cellcolor{lightorange!15}28.49 \\
history & 52.30 & 34.20 & \cellcolor{lightred!15}53.45 & 36.48 & 49.43 & 35.10 & 50.00 & 34.51 & 52.30 & \cellcolor{lightorange!15}29.80 \\
law & 29.40 & 39.76 & 30.15 & 45.61 & \cellcolor{lightred!15}31.84 & 45.29 & 30.71 & 48.24 & 31.08 & \cellcolor{lightorange!15}37.58 \\
math & 37.18 & 36.68 & 37.33 & 30.98 & 35.81 & 33.23 & 36.42 & 31.83 & \cellcolor{lightred!15}39.00 & \cellcolor{lightorange!15}27.19 \\
philosophy & 45.49 & 43.53 & \cellcolor{lightred!15}46.78 & 36.62 & 45.92 & 41.23 & 44.64 & 43.20 & \cellcolor{lightred!15}46.78 & \cellcolor{lightorange!15}33.84 \\
physics & 41.23 & 36.59 & 40.60 & 28.77 & 41.71 & 27.46 & 40.13 & 33.15 & \cellcolor{lightred!15}42.97 & \cellcolor{lightorange!15}25.78 \\
psychology & 65.45 & 33.09 & 69.37 & 27.56 & 68.32 & 28.75 & 67.28 & 31.78 & \cellcolor{lightred!15}69.89 & \cellcolor{lightorange!15}25.83 \\
\midrule
Average  & 43.48 & 31.73 & 44.55 & 28.06 & 43.62 & 29.11 & 43.32 & 30.87 & \cellcolor{lightred!15}45.31 & \cellcolor{lightorange!15}25.24  \\
\bottomrule
\end{tabular}%
\caption{The test accuracy and ECE of optimization-free approaches on MMLU-Pro dataset with \texttt{Qwen3-8B}. While sometimes simply using all exemplars or setting up a retrieval system for online search obtains a better accuracy, our calibration-aware search outperforms all baselines with respect to ECE, establishing confidence in LLM predictions. The results for \texttt{LLaMA-3.3-70B} can be found in App.~\ref{sec:more_results}.}
\label{tab:opt_free_testset_qwen}
\end{table*}

\begin{figure*}[htbp]
  \centering
    \includegraphics[width=1.0\linewidth]{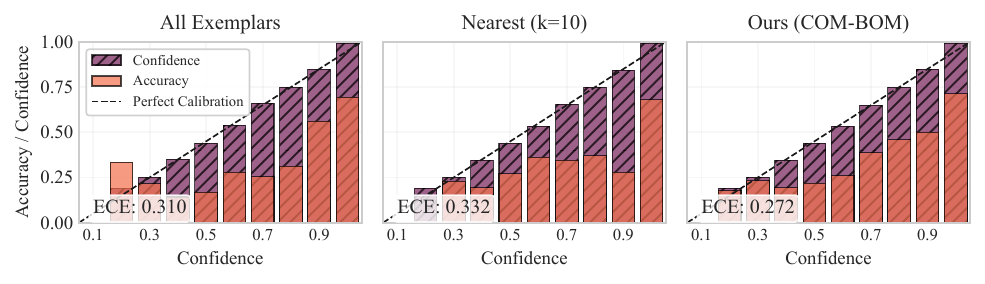}
  \caption{The reliability diagram of test accuracy and ECE for \texttt{Math} task from MMLU-Pro. It highlights the necessity of optimizing for calibration error that is paramount for deploying trustworthy LLM systems, in order to minimize over-confident wrong predictions and under-confident right predictions. Compared to online retrieval systems for exemplar search, \ours is more cost-effective at inference time due to its offline search.}
  \label{fig:reliability_math}
\end{figure*}

\begin{table*}[htbp]
\centering
\small
\begin{tabular}{l|cc|cc|cc|cc|cc}
\toprule
& \multicolumn{2}{c|}{Genetic Algorithm} & \multicolumn{2}{c|}{Simulated Annealing} & \multicolumn{2}{c|}{Random Search} & \multicolumn{2}{c|}{Hill Climbing} & \multicolumn{2}{c}{Ours} \\
\cmidrule(lr){2-3} \cmidrule(lr){4-5} \cmidrule(lr){6-7} \cmidrule(lr){8-9} \cmidrule(lr){10-11}
Domain & Acc & ECE & Acc & ECE & Acc & ECE & Acc & ECE & Acc & ECE \\
\midrule
biology & 76.61 & 20.37 & 77.49 & 22.90 & 78.65 & 20.32 & 77.19 & 20.00 & \cellcolor{lightred!15}79.24 & \cellcolor{lightorange!15}18.37 \\
business & 37.83 & 23.89 & 38.10 & 23.13 & 37.57 & 33.44 & \cellcolor{lightred!15}41.53 & 28.26 & 41.01 & \cellcolor{lightorange!15}21.54 \\
chemistry & 34.06 & 29.37 & 36.07 & 30.18 & 34.24 & 30.41 & 35.70 & 33.04 & \cellcolor{lightred!15}37.34 & \cellcolor{lightorange!15}28.38 \\
computer science & 48.94 & 38.71 & 51.06 & 34.71 & 48.94 & 36.44 & 51.60 & 35.62 & \cellcolor{lightred!15}55.85 & \cellcolor{lightorange!15}32.08 \\
economics & 62.22 & 31.29 & 62.72 & 31.24 & 62.22 & 31.75 & 62.72 & \cellcolor{lightorange!15}30.35 & \cellcolor{lightred!15}63.70 & 30.74 \\
engineering & 45.51 & 21.42 & 44.23 & 22.57 & 43.80 & 20.22 & \cellcolor{lightred!15}48.50 & 24.72 & 47.01 & \cellcolor{lightorange!15}19.32  \\
health & 61.22 & 32.79 & 58.67 & 32.78 & 60.20 & 34.84 & 59.95 & 29.11 & \cellcolor{lightred!15}62.24 & \cellcolor{lightorange!15}28.49 \\
history & \cellcolor{lightred!15}52.87 & \cellcolor{lightorange!15}23.72 & 47.70 & 30.73 & 51.15 & 38.74 & 51.15 & 27.49 & 52.30 & 29.80  \\
law & 31.09 & 38.68 & 31.46 & 42.58 & \cellcolor{lightred!15}32.58 & 44.95 & 31.84 & 39.10 & 31.08 & \cellcolor{lightorange!15}37.58 \\
math & 36.87 & 33.92 & 37.02 & 28.48 & 37.32 & 31.51 & 37.48 & 30.23 & \cellcolor{lightred!15}39.00 & \cellcolor{lightorange!15}27.19  \\
philosophy & 45.06 & 39.28 & 44.63 & 41.41 & 44.21 & 40.01 & 43.34 & 39.12 & \cellcolor{lightred!15}46.78 & \cellcolor{lightorange!15}33.84 \\
physics & 40.28 & 31.32 & 39.34 & 29.33 & 40.13 & 33.20 & 40.76 & 29.19 & \cellcolor{lightred!15}42.97 & \cellcolor{lightorange!15}25.78 \\
psychology & 67.28 & 28.02 & 67.80 & 29.33 & 68.06 & 31.51 & 68.84 & \cellcolor{lightorange!15}24.94 & \cellcolor{lightred!15}69.89 & 25.83  \\
\midrule
Average & 43.36 & 27.98 & 43.26 & 28.02 & 43.41 & 29.92 & 44.24 & 27.73 & \cellcolor{lightred!15}45.31 & \cellcolor{lightorange!15}25.24  \\
\bottomrule
\end{tabular}%
\caption{The test accuracy and ECE of optimization-based approaches on MMLU-Pro dataset with \texttt{Qwen3-8B}. The results for \texttt{LLaMA-3.3-70B} can be found in Appendix \ref{sec:more_results}.}
\label{tab:opt_based_testset_qwen}
\end{table*}

\paragraph{Models and datasets.}
We evaluate \ours using Qwen models (\texttt{Qwen3-8B})~\cite{qwen3} and LLaMA models (\texttt{LlaMA-3.3-70B})~\cite{llama3} on an extensive collection of tasks from MMLU-Pro~\cite{mmlupro}, a challenging benchmark where LLMs have not yet achieved performance saturation (as of May 2025). In addition, MMLU-Pro covers a diverse set of tasks beyond math and coding problems where data contamination is a known issue~\cite{matharena}. This makes it a suitable benchmark for evaluating methods that leverage exemplars to 
improve few-shot learning of LLMs on novel tasks, without requiring fine-tuning. For each task, we construct an exemplar pool by randomly selecting 32 samples, with the remaining samples divided equally between validation and test, the latter of which is held-out and unavailable to LLMs at search time. We refer readers to App.~\ref{sec:impl_details} for implementation details, including prompt template and sampling parameters.

\paragraph{Experimental setup and baselines.}
We benchmark \ours against multiple optimization-based techniques, including random search (RS), genetic algorithm (GA), simulated annealing (SA), and hill climbing (HC) with scalarization, because they represent standard approaches, frequently employed in practice, for combinatorial black-box optimization problems~\cite{deshwal2021mercer,mcbo}. Following \citet{teach_or_show}, we also evaluate against optimization-free retrieval baselines: Nearest and Diversity. The Nearest retrieves the $k$ exemplars with highest text-embedding cosine similarity to the input query, while Diversity selects $k$ input-output pairs closest to centroids identified through $k$-means clustering in the embedding space~\cite{diversity}. For the text-embedding model, we used \texttt{stella\_en\_400M\_v5}~\cite{zhang2024jasper-dbd} as it achieved the highest overall score among lightweight models (<1B parameters) on MTEB benchmark~\cite{mmteb}, as of May 2025. Additionally, we also include the two simplest baselines: using no exemplars and using all the exemplars. 

\paragraph{Results and discussion.}
Nearly all tasks show improvement from ICL with exemplars (Tab.~\ref{tab:opt_free_testset_qwen}) as \texttt{Qwen3} is pretrained on 36T tokens spanning diverse domains including (non-)STEM fields. These supervised exemplars effectively focus the model's pre-training distribution toward domain-specific parametric knowledge~\cite{icl_theory}. However, simply increasing number of exemplars does not guarantee monotonically increasing performance \cite{agarwal2024many}, as individual exemplars contribute differently to task outcomes, emphasizing the importance of strategic exemplar selection. 

\begin{figure*}[htbp]
  \centering
    \includegraphics[width=1.0\linewidth]{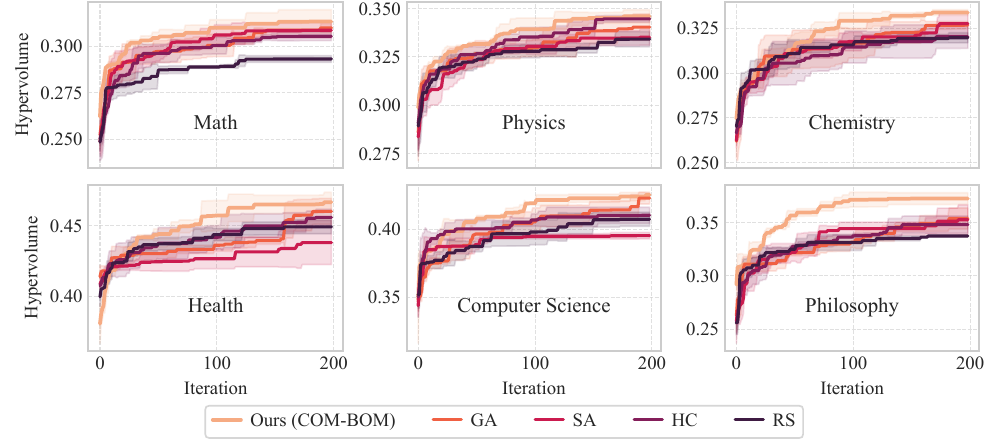}
  \caption{Evolution of best observed hypervolume on the validation data across \texttt{STEM}, \texttt{Medical} and \texttt{Humanity} tasks for optimization approaches. The hypervolume is measured against the reference point (accuracy=0\%, ECE=100\%). The evolution plots the average of three runs. Please see App.~\ref{sec:more_results} for results on rest of the tasks.}
  \label{fig:opt_valset}
  \vspace{-1em}
\end{figure*}

Our results reveal that online retrieval-based approaches underperform compared to \ours which is an offline (which only leverage a fixed validation set) search method (Tab.~\ref{tab:opt_free_testset_qwen} and Fig.~\ref{fig:reliability_math}). This finding persists across various values of $k$ in the retrieval system (App.~\ref{sec:more_results}). While offline search has initial computational costs, these can be amortized during deployment, yielding greater cost-effectiveness at inference time. Additionally, we observe consistently lower performance with diversity-based exemplar selection, likely due to the inherent difficulty of clustering in the high-dimensional  space.

\begin{figure}[htbp]
    \centering
    \includegraphics[width=1.0\linewidth]{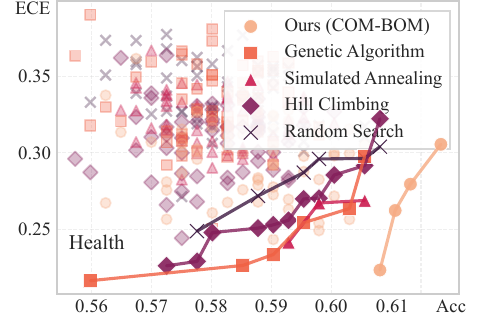}
    \caption{The observed validation accuracy and ECE with Pareto frontiers on \texttt{Health} task for optimization-based approaches with fixed evaluation iterations. \ours identifies better Pareto-optimal solutions.}
    \label{fig:pareto_fronts}
\end{figure}

Our results also demonstrate that optimization approaches consistently outperform optimization-free baselines (Tab.~\ref{tab:opt_based_testset_qwen}), with \ours exhibiting superior sample efficiency in solving the multi-objective optimization problem (Fig.~\ref{fig:opt_valset}). This efficiency translates to an improved accuracy-calibration trade-off while requiring fewer LLM API calls. Notably, \ours offers a distinct advantage over other  optimization baselines, that rely on scalarization, as it directly reasons about the Pareto front in terms of its hypervolume. Consequently, \ours shows faster convergence toward Pareto-optimal solutions within a small number of evaluation iterations (Fig.~\ref{fig:pareto_fronts}).

\begin{figure}[htbp]
    \centering
    \includegraphics[width=1.0\linewidth]{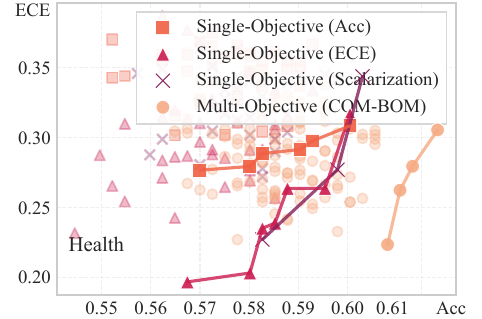}
    \caption{The observed validation accuracy and ECE with Pareto frontiers on \texttt{Health} task for single versus multi-objective formulation. Interestingly, optimizing for calibration error yields better Pareto frontiers than optimizing for accuracy alone. \ours leverages both objectives more effectively than scalarization.}
    \label{fig:ablation_single_objective_valset}
\end{figure}

\paragraph{Necessity of MOO formulation.} To our knowledge,  we are the first to propose calibration-aware exemplar selection as a multi-objective optimization formulation. To demonstrate the importance of such formulation, we also compare with a single-objective combinatorial BO baseline. Our results demonstrate that optimizing for accuracy alone often compromises calibration, resulting in predictors with sub-optimal reliability (Fig.~\ref{fig:ablation_single_objective_valset}). Conversely optimizing exclusively for ECE compromises accuracy, yielding less effective predictors. While scalarization provides some benefit, \ours finds much better Pareto frontiers of the two objectives.

\begin{figure}[htbp]
    \centering
    \includegraphics[width=1.0\linewidth]{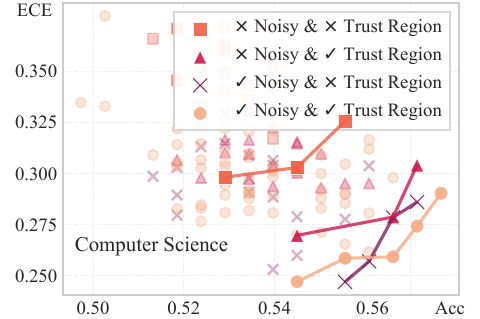}
    \caption{The observed validation accuracy and ECE with Pareto frontiers on \texttt{Computer} \texttt{Science} task if accounting for noisy observation and trust region.}
    \label{fig:ablation_study_bo}
\end{figure}

\paragraph{Ablative study of BO components.} Our ablation analysis (Fig.~\ref{fig:ablation_study_bo}) demonstrates that incorporating both noisy observations \cite{daulton2021parallel} and trust region \cite{eriksson2019scalable} yields the most favorable trade-offs, achieving lower ECE at higher accuracy. Removing either or both components results in degraded performance highlighting the complementary benefits of local search and handling noisy observations.

\section{Conclusion}
We introduce a new formulation of exemplar selection as a multi-objective optimization problem where the key idea is to ensure reliability by jointly optimizing for calibration and accuracy metrics  rather than optimizing the latter alone. Additionally, we propose a sample-efficient combinatorial Bayesian optimization algorithm \ours for this black-box optimization problem that involves multiple objectives and is expensive to evaluate. Extensive experiments show that \ours achieves better Pareto front with fewer LLM evaluations compared to existing methods.

\paragraph{Future work.} There are multiple avenues for future work in this problem space. In our Bayesian optimization approach, we used a Gaussian process model with Exponentiated Hamming Kernel as the surrogate model, a simple yet effective choice for our search space. For even larger search spaces, this kernel can be replaced with more sophisticated ones that operate on learned embeddings for exemplars, allowing the model to leverage semantic similarity and scale more effectively. Another fertile ground for exploration is to consider the additional input space of the order in which exemplars are included in the prompt. Recent work in Bayesian optimization over permutation spaces \cite{deshwal2022bayesian} can be an effective direction in order to tackle such  an input space of exemplar orderings.

\section{Limitations} Our experiments are based on an exemplar pool of 32 candidates. We consider this pool size to be practical for novel tasks where only a limited number of high-quality, human-annotated demonstrations can be sourced. Furthermore, using a large number of exemplars during inference can introduce undesirable latency and serving costs for diminishing gains in performance. However, Bayesian Optimization is known to face challenges with scalability as the dimensionality of the search space (in our case, number of candidate exemplars) increases. 
Due to computational resource constraints, our evaluations were limited to dense LLMs with fewer than <100B parameters. We specifically choose \texttt{Qwen3-8B} and \texttt{LLaMA-3.3-70B} since they are pre-trained on the largest corpora among models of the same size to elicit in-context learning behaviors.  
Additionally, our evaluation was restricted to English-language multiple-choice QA tasks from the MMLU-Pro dataset.  Expanding this methodology to open-ended reasoning tasks (e.g., code generation) would be a valuable extension, though it would require more nuanced metrics for quantifying accuracy (e.g., pass@k) and confidence (e.g., semantic uncertainty). To ensure consistent output formatting, especially with models of this scale, we reserved one exemplar exclusively for demonstrating the desired format. We empirically observe that the output format is almost always enforced for multiple-choice QA tasks employed in our paper.

\section{Ethics Statement}
We are using open-source models (\texttt{Qwen3-8B} and \texttt{LLaMA-3.3-70B}) and data (MMLU-Pro). This work essentially modifies the input to an LLM by selecting better exemplars to include in the few-shot prompt, and thus does not introduce additional risk to LLMs themselves. Instead, this work implicitly finds exemplars that cause an LLM to be faithful to its own confidence in its prediction, making it more reliable at deployment. However, if an adversary collects a pool of popular malicious prompts such as jailbreaking exemplars, with a well-defined performance metric such as success rate, our method can potentially be used to find the Pareto-optimal sets of malicious exemplars that maximize the success rate and minimize the calibration errors (i.e., confidently break the system) with few black-box LLM evaluations only.

\section*{Acknowledgment}

This work is supported by NSF grant IIS-2313174 and a Data Science Graduate Assistantship with funding made available by the MnDRIVE initiative through the University of Minnesota Data Science Initiative. The authors acknowledge the Minnesota Supercomputing Institute (MSI) at the University of Minnesota for providing resources that contributed to the results reported within this paper.

\bibliography{acl_latex}

\clearpage

\appendix

\section{Implementation Details} \label{sec:impl_details}

\begin{tcolorbox}[colback=black!5!white,colframe=black!75!black,title=MMLU-Pro Prompt Template]
The following are multiple choice questions (with examples) about \{\{ DOMAIN \}\}. When you provide the answer to the last question, please use the option letter without any modification, and provide the answer directly, with no formatting, no bolding, and no markup. For example, (A). The final answer must only be the letter corresponding to the correct answer. \\

\{\{ EXEMPLARS \}\} \\

Question:	\\
\{\{ QUESTION \}\} \\
Options: \\
\begin{tabular}{@{}ll}
A. & \{\{ OPTION\_A \}\} \\
B. & \{\{ OPTION\_B \}\} \\
C. & \{\{ OPTION\_C \}\} \\
D. & \{\{ OPTION\_D \}\} \\
E. & \{\{ OPTION\_E \}\} \\
F. & \{\{ OPTION\_F \}\} \\
G. & \{\{ OPTION\_G \}\} \\
H. & \{\{ OPTION\_H \}\} \\
I. & \{\{ OPTION\_I \}\} \\
J. & \{\{ OPTION\_J \}\} \\
\end{tabular}\\
The answer is: 
\end{tcolorbox}

\begin{table}[htbp]
    \centering
    \begin{tabular}{cc}
    \toprule
    \multicolumn{2}{c}{Hyper-parameters}  \\
    \midrule
    \# Exemplars &  32 \\
    LLM 1 & Qwen3-8B \\
    LLM 2 & LLaMA-3.3-70B-Instruct \\
    Temperature & 0.7 \\
    \# Samples & 16 \\
    Top P & 0.8 \\
    Top K & 20 \\
    Eval Budget & 200 Iterations \\
    \# BO Initial Points & 20 \\
    Trust Region & Local \\
    Acquisition Function & qNEHVI \\
    Seeds & 0,1,2 \\
    \bottomrule
    \end{tabular}
    \caption{Implementation Details of MMLU-Pro Experiments.}
    \label{tab:impl_mmlupro}
\end{table}

The MMLU-Pro dataset~\cite{mmlupro} is licensed under the MIT License, \texttt{Qwen-8B} model~\cite{qwen3} is publicly accessible under Apache 2.0 and \texttt{LLaMA-3.3-70B} model ~\cite{llama3} is licensed under their own community license agreement. The models and dataset used in this paper are for research purposes only. The dataset does not contain personal identifiable information or offensive content. The dataset consists of complex questions in English in various disciplines, including the domains described in Tab.~\ref{tab:opt_based_testset_k}, where each question has at most 10 options. The size of the validation set has to be reasonably large to obtain a distinguishable accuracy and ECE over search iterations. Specifically, for validation we use 174 samples in \texttt{History}, 188 samples in \texttt{Computer} \texttt{Science}, 233 samples in \texttt{Philosophy}, 342 samples in \texttt{Biology}, 378 samples in \texttt{Business}, 382 samples in \texttt{Psychology}, 392 samples in \texttt{Health}, 485 samples in \texttt{Economics}, 468 samples in \texttt{Engineering}, 534 samples in \texttt{Law}, 549 samples in \texttt{Chemistry}, 633 samples in \texttt{Physics} and 659 samples in \texttt{Math}. In addition, we have the same number of non-overlapping test samples for each task. In addition to validation and test data, we have 32 samples for the exemplar pool and 1 sample set aside for output formatting.

All \texttt{Qwen3-8B} experiments were performed on 2x NVIDIA A100-SXM4-40GB because an 8B model in full precision roughly occupies 56GB GPU memory. The total number of runtime is approximately 126 GPU hours for all \texttt{Qwen3-8B} experiments. All optimization-based methods are repeated 3 times with 3 different random seeds on the search, and non-dominated Pareto frontier points are used to evaluate on the test data respectively. All optimization-free baselines are repeated 3 times before recording the mean on the test data.


\section{More Experimental Results}\label{sec:more_results}
To demonstrate the generality of \ours, we conducted additional experiments on \texttt{LLaMA-3.3-70B}, a representative larger LLM. Similar to our experiments with \texttt{Qwen3-8B}, using all exemplars does not necessarily lead to optimal performance, indicating that exemplar selection is critical. Among all the optimization-based and optimization-free methods, \ours consistently finds an optimal exemplar set that maximizes accuracy while minimizing calibration error. 
We find online retrieval-based methods can sometimes be competitive in accuracy but consistently compromise calibration error. In contrast, through its multi-objective formulation, \ours aims to identify exemplars with optimal tradeoff between accuracy and  calibration error.

\begin{table*}[htbp]
\centering
\small
\begin{tabular}{l|cc|cc|cc|cc|cc}
\toprule
& \multicolumn{2}{c|}{No Exemplars} & \multicolumn{2}{c|}{All Exemplars} & \multicolumn{2}{c|}{$\text{Nearest}_{k=10}$} & \multicolumn{2}{c|}{$\text{Diversity}_{k=10}$} & \multicolumn{2}{c}{Ours} \\
\cmidrule(lr){2-3} \cmidrule(lr){4-5} \cmidrule(lr){6-7} \cmidrule(lr){8-9} \cmidrule(lr){10-11}
Domain & Acc & ECE & Acc & ECE & Acc & ECE & Acc & ECE & Acc & ECE \\
\midrule
biology & \cellcolor{lightred!15}81.29 & 14.94 & 78.65 & 16.48 & 79.23 & 16.94 & 78.36 & 19.05 & 79.23 & \cellcolor{lightorange!15}13.37 \\
business & 38.62 & 29.99 & 40.21 & 31.87 & 40.74 & 32.35 & 42.86 & 33.00 & \cellcolor{lightred!15}47.09 & \cellcolor{lightorange!15}29.87 \\
chemistry & 35.52 & 38.53 & 42.26 & 32.95 & 41.35 & 35.72 & 42.62 & 37.35 & \cellcolor{lightred!15}47.18 & \cellcolor{lightorange!15}30.74 \\
computer science & 47.87 & \cellcolor{lightorange!15}19.14 & 55.32 & 30.68 & 56.91 & 32.18 & 52.66 & 29.98 & \cellcolor{lightred!15}57.45 & 26.73 \\
economics & 67.16 & 23.16 & 67.16 & 24.53 & 64.44 & 23.90 & 64.69 & 25.34 & \cellcolor{lightred!15}67.40 & \cellcolor{lightorange!15}18.52 \\
engineering & 37.82 & 29.88 & 35.68 & 28.06 & 39.10 & 29.90 & 35.89 & 30.69 &\cellcolor{lightred!15}40.81 & \cellcolor{lightorange!15}27.18 \\
health & 71.17 & 24.04 & 69.64 & 24.67 & 68.62 & 25.12 & 70.66 & 26.16 & \cellcolor{lightred!15}71.17 & \cellcolor{lightorange!15}22.56 \\
history & 70.69 & \cellcolor{lightorange!15}20.75 & 67.82 & 24.39 & 70.11 & 22.08 & 68.96 & 24.25 & \cellcolor{lightred!15}71.26 & 22.95 \\
law & 50.56 & 41.46 & 51.31 & 40.02 & \cellcolor{lightred!15}53.74 & 38.88 & 51.68 & 40.70 &52.24 & \cellcolor{lightorange!15}36.24 \\
math & 27.16 & 39.59 & 35.36 & 36.53 & 31.87 & 40.68 & 31.41 & 39.58 & \cellcolor{lightred!15}37.18 & \cellcolor{lightorange!15}34.37 \\
philosophy & 62.23 & 27.79 & 57.51 & 30.58 & 62.66 & 29.82 & 60.01 & 32.38 & \cellcolor{lightred!15}64.81 & \cellcolor{lightorange!15}25.92 \\
physics & 36.49 & 39.55 & 45.50 & 36.12 & 43.75 & 35.59 & 41.70 & 40.06 &\cellcolor{lightred!15}47.86 & \cellcolor{lightorange!15}32.47 \\
psychology & 73.56 & 22.70 & 78.01 & 20.17 & \cellcolor{lightred!15}78.27 & 20.48 & 76.96 & 20.83 & 77.25 & \cellcolor{lightorange!15}18.09 \\
\midrule
Average  & 46.10 & 28.71 & 48.66 & 27.95 & 48.64 & 28.69 & 47.91 & 29.90 & \cellcolor{lightred!15}51.12 & \cellcolor{lightorange!15}25.30 \\
\bottomrule
\end{tabular}%
\caption{The test accuracy and ECE of optimization-free approaches on MMLU-Pro dataset with \texttt{LLaMA3-3.3-70B}.}
\label{tab:opt_free_testset_llama}
\end{table*}

\begin{table*}[htbp]
\centering
\small
\begin{tabular}{l|cc|cc|cc|cc|cc}
\toprule
& \multicolumn{2}{c|}{Genetic Algorithm} & \multicolumn{2}{c|}{Simulated Annealing} & \multicolumn{2}{c|}{Random Search} & \multicolumn{2}{c|}{Hill Climbing} & \multicolumn{2}{c}{Ours} \\
\cmidrule(lr){2-3} \cmidrule(lr){4-5} \cmidrule(lr){6-7} \cmidrule(lr){8-9} \cmidrule(lr){10-11}
Domain & Acc & ECE & Acc & ECE & Acc & ECE & Acc & ECE & Acc & ECE \\
\midrule
biology & \cellcolor{lightred!15}81.28 & 15.64 & 79.53 & 17.56 & 80.11 & 19.20 & 80.11 & 15.54 & 79.23 & \cellcolor{lightorange!15}13.37 \\
business & 43.12 & 32.52 & 44.70 & 32.23 & 42.33 & 32.42 & 43.12 & 31.69 & \cellcolor{lightred!15}47.09 & \cellcolor{lightorange!15}29.87 \\
chemistry & 44.08 & 32.54 & 44.08 & 34.32 & 45.53 & 31.99 & 45.17 & 30.97 & \cellcolor{lightred!15}47.18 & \cellcolor{lightorange!15}30.74 \\
computer science & 55.85 & 29.53 & 56.91 & 30.60 & 52.66 & 31.26 & 54.79 & 28.01 & \cellcolor{lightred!15}57.45 & \cellcolor{lightorange!15}26.73 \\
economics & \cellcolor{lightred!15}67.65 & 21.60 & 66.66 & 21.23 & 66.91 & 25.58 & 65.67 & 24.20 & 67.40 & \cellcolor{lightorange!15}18.52 \\
engineering & 39.31 & 29.00 & 36.75 & 32.12 & 36.11 & 33.93 & 39.95 & 29.74 & \cellcolor{lightred!15}40.81 & \cellcolor{lightorange!15}27.18 \\
health & 69.38 & 22.39 & 68.36 & \cellcolor{lightorange!15}19.92 & 69.64 & 26.13 & 68.88 & 23.23 & \cellcolor{lightred!15}71.17 & 22.56 \\
history & 69.54 & 28.08 & 68.39 & 25.84 & 68.96 & 22.70 & 70.11 & 26.22 & \cellcolor{lightred!15}71.26 & \cellcolor{lightorange!15}22.95 \\
law & 51.87 & 39.50 & 51.68 & 37.88 & 51.68 & 41.65 & 52.05 & 38.80 & \cellcolor{lightred!15}52.24 & \cellcolor{lightorange!15}36.24 \\
math & 33.54 & 37.80 & 34.90 & 39.85 & 35.66 & 38.08 & 34.46 & 40.04 & \cellcolor{lightred!15}37.18 & \cellcolor{lightorange!15}34.37 \\
philosophy & 59.23 & 26.01 & 59.66 & 26.55 & 58.80 & 30.35 & 61.80 & 26.54 & \cellcolor{lightred!15}64.81 & \cellcolor{lightorange!15}25.92 \\
physics & 45.18 & 34.61 & 45.33 & 37.23 & 42.33 & 39.35 & 45.33 & 34.07 & \cellcolor{lightred!15}47.86 & \cellcolor{lightorange!15}32.47 \\
psychology & 76.43 & 18.90 & 75.13 & 24.30 & \cellcolor{lightred!15}77.74 & 22.01 & 75.39 & 19.68 & 77.25 & \cellcolor{lightorange!15}18.09 \\
\midrule
Average & 49.34 & 27.36 & 49.07 & 28.40 & 48.92 & 29.47 & 49.40 & 27.55 & \cellcolor{lightred!15}51.12 & \cellcolor{lightorange!15}25.30 \\
\bottomrule
\end{tabular}%
\caption{The test accuracy and ECE of optimization-based approaches on MMLU-Pro dataset with \texttt{LLaMA-3.3-70B}.}
\label{tab:opt_based_testset_llama}
\end{table*}

\begin{table*}[htbp]
\centering
\small
\begin{tabular}{l|cc|cc|cc|cc}
\toprule
& \multicolumn{2}{c|}{$\text{Nearest}_{k=5}$} & \multicolumn{2}{c|}{$\text{Nearest}_{k=10}$} & \multicolumn{2}{c|}{$\text{Nearest}_{k=20}$} & \multicolumn{2}{c}{Ours} \\
\cmidrule(lr){2-3} \cmidrule(lr){4-5} \cmidrule(lr){6-7} \cmidrule(lr){8-9}
Domain & Acc & ECE & Acc & ECE & Acc & ECE & Acc & ECE  \\
\midrule
biology & 77.19 & 20.47 & 76.60 & 20.29 & 78.95 & 19.11 & \cellcolor{lightred!15}79.24 & \cellcolor{lightorange!15}18.37  \\
business & \cellcolor{lightred!15}42.06 & 29.53 & 39.68 & 30.72 & 39.95 & 30.15 & 41.01 & \cellcolor{lightorange!15}21.54  \\
chemistry & 36.79 & 33.28 & \cellcolor{lightred!15}37.34 & 31.79 & 36.43 & 31.61 & \cellcolor{lightred!15}37.34 & \cellcolor{lightorange!15}28.38  \\
computer science & 51.06 & 36.21 & 48.94 & 37.16 & 51.60 & 34.62 & \cellcolor{lightred!15}55.85 & \cellcolor{lightorange!15}32.08  \\
economics & \cellcolor{lightred!15}63.70 & 31.91 & 61.48 & 30.98 & 62.47 & \cellcolor{lightorange!15}29.10 & \cellcolor{lightred!15}63.70 & 30.74  \\
engineering & \cellcolor{lightred!15}48.08 & 20.97 & 47.01 & 21.00  & 46.79 & 19.71  & 47.01 & \cellcolor{lightorange!15}19.32   \\
health & 59.18 & 30.21 & 56.12 & 32.55 & 59.69 & 30.20 & \cellcolor{lightred!15}62.24 & \cellcolor{lightorange!15}28.49  \\
history & \cellcolor{lightred!15}52.87 & 34.43 & 49.43 & 35.10 & 50.57 & 34.46 & 52.30 & \cellcolor{lightorange!15}29.80   \\
law & \cellcolor{lightred!15}32.96 & 42.96 & 31.84 & 45.29 & 29.78 & 45.65 & 31.08 & \cellcolor{lightorange!15}37.58   \\
math & 36.27 & 34.48 & 35.81 & 33.23 & 37.48 & 30.99 & \cellcolor{lightred!15}39.00 & \cellcolor{lightorange!15}27.19   \\
philosophy & 44.64 & 42.73 & 45.92 & 41.23 & 43.35 & 42.61 & \cellcolor{lightred!15}46.78 & \cellcolor{lightorange!15}33.84  \\
physics & 39.81 & 30.83 & 41.71 & 27.46 & 39.49 & 27.39 & \cellcolor{lightred!15}42.97 & \cellcolor{lightorange!15}25.78  \\
psychology & 67.80 & 27.23 & 68.32 & 28.75 & 69.63 & 27.44 & \cellcolor{lightred!15}69.89 & \cellcolor{lightorange!15}25.83   \\
\midrule
Average & 44.24 & 29.29 & 43.62 & 29.11 & 43.84 & 28.22 & \cellcolor{lightred!15}45.31 & \cellcolor{lightorange!15}25.24  \\
\bottomrule
\end{tabular}%
\caption{The test accuracy and ECE of online retrieval-based approach on MMLU-Pro dataset with \texttt{Qwen3-8B}. With $k\in\{5,10,20\}$, \ours demonstrates superior performance (especially ECE) while being cost-effective at inference time due to its offline search before deployment with only a fixed validation set.}
\label{tab:opt_based_testset_k}
\end{table*}

\begin{figure*}[t]
  \centering
    \includegraphics[width=1.0\linewidth]{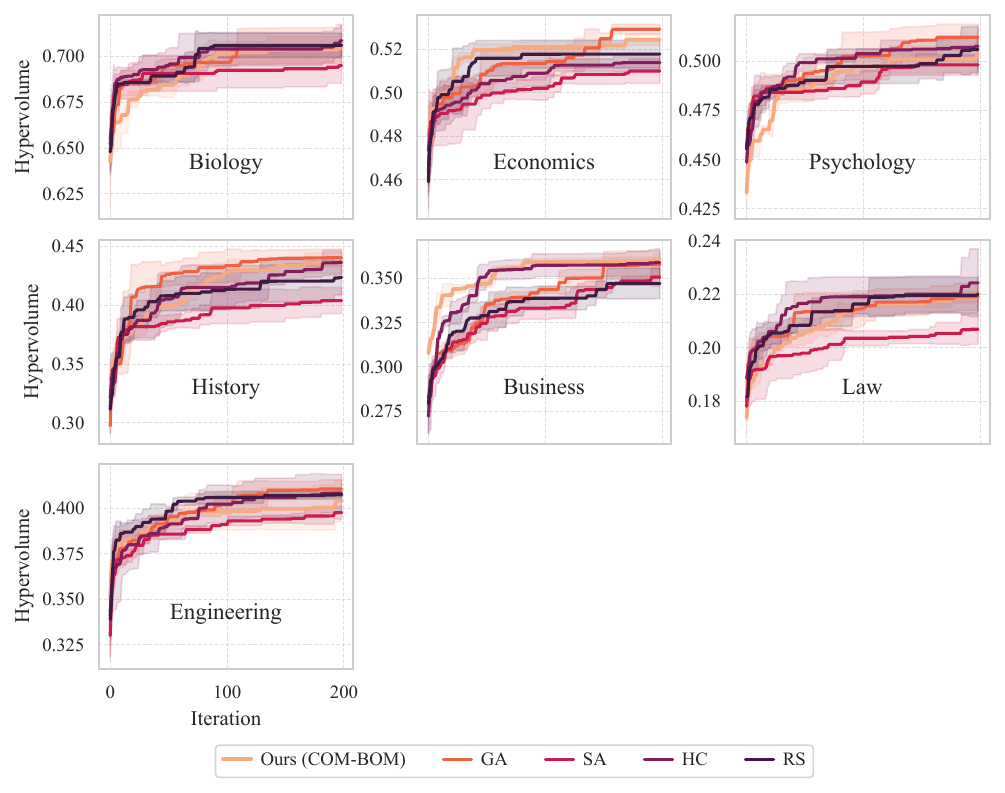}
  \caption{Evolution of best observed hypervolume on the validation data across the remaining MMLU-Pro tasks for optimization approaches. The hypervolume is measured against the reference point (accuracy=0\%, ECE=100\%).}
  \label{fig:opt_valset_cont}
\end{figure*}

\end{document}